\documentclass{article}

\usepackage{microtype}
\usepackage{graphicx}
\usepackage{subfigure}
\usepackage{booktabs} % for professional tables

\usepackage[most]{tcolorbox}
\usepackage{hyperref}

\usepackage[accepted]{icml2025}

\usepackage{amsmath}
\usepackage{amssymb}
\usepackage{mathtools}
\usepackage{amsthm}

\usepackage[capitalize,noabbrev]{cleveref}

\theoremstyle{plain}

\theoremstyle{definition}

\theoremstyle{remark}

\usepackage[textsize=tiny]{todonotes}

\newcommand{\systemname}[0]{\textsc{ProofCompass}}

\icmltitlerunning{\systemname{}: Enhancing Specialized Provers with LLM Guidance}

\begin{document}

\twocolumn[
\icmltitle{\systemname{}: Enhancing Specialized Provers with LLM Guidance}



\icmlsetsymbol{equal}{*}

\begin{icmlauthorlist}
\icmlauthor{Nicolas Wischermann}{xxx}
\icmlauthor{Claudio Mayrink Verdun}{yyy}
\icmlauthor{Gabriel Poesia}{zzz}
\icmlauthor{Francesco Noseda}{xxx}

\end{icmlauthorlist}

\icmlaffiliation{xxx}{Federal University of Rio de Janeiro}
\icmlaffiliation{yyy}{Harvard John A. Paulson School of Engineering and Applied Sciences}
\icmlaffiliation{zzz}{Stanford University}

\icmlcorrespondingauthor{Nicolas Wischermann}{ndasilva@ufrj.br}

\icmlkeywords{Machine Learning, ICML}

\vskip 0.3in
]



\printAffiliationsAndNotice{}  % leave blank if no need to mention equal contribution

\begin{abstract}
Language models have become increasingly powerful tools for formal mathematical reasoning. However, most existing approaches rely exclusively on either large general-purpose models or smaller specialized models, each with distinct limitations, while training specialized large models still requires significant computational resources. This paper introduces \systemname{}, a novel hybrid methodology that achieves remarkable computational efficiency by strategically guiding existing specialized prover methods, such as DeepSeek-Prover-v1.5-RL (DSP-v1.5) with a Large Language Model (LLM) without requiring additional model training. The LLM provides natural language proof strategies and analyzes failed attempts to select intermediate lemmas, enabling effective problem decomposition. On the miniF2F benchmark, \systemname{} demonstrates substantial resource efficiency: it outperforms DSP-v1.5 ($54.9\% \rightarrow 55.3\%$) while using 25x fewer attempts ($3200 \rightarrow 128$). Our synergistic approach paves the way for simultaneously improving computational efficiency and accuracy in formal theorem proving.
\end{abstract}

\section{Introduction}

\begin{figure}[ht]
  \centering
  \includegraphics[width=\columnwidth]{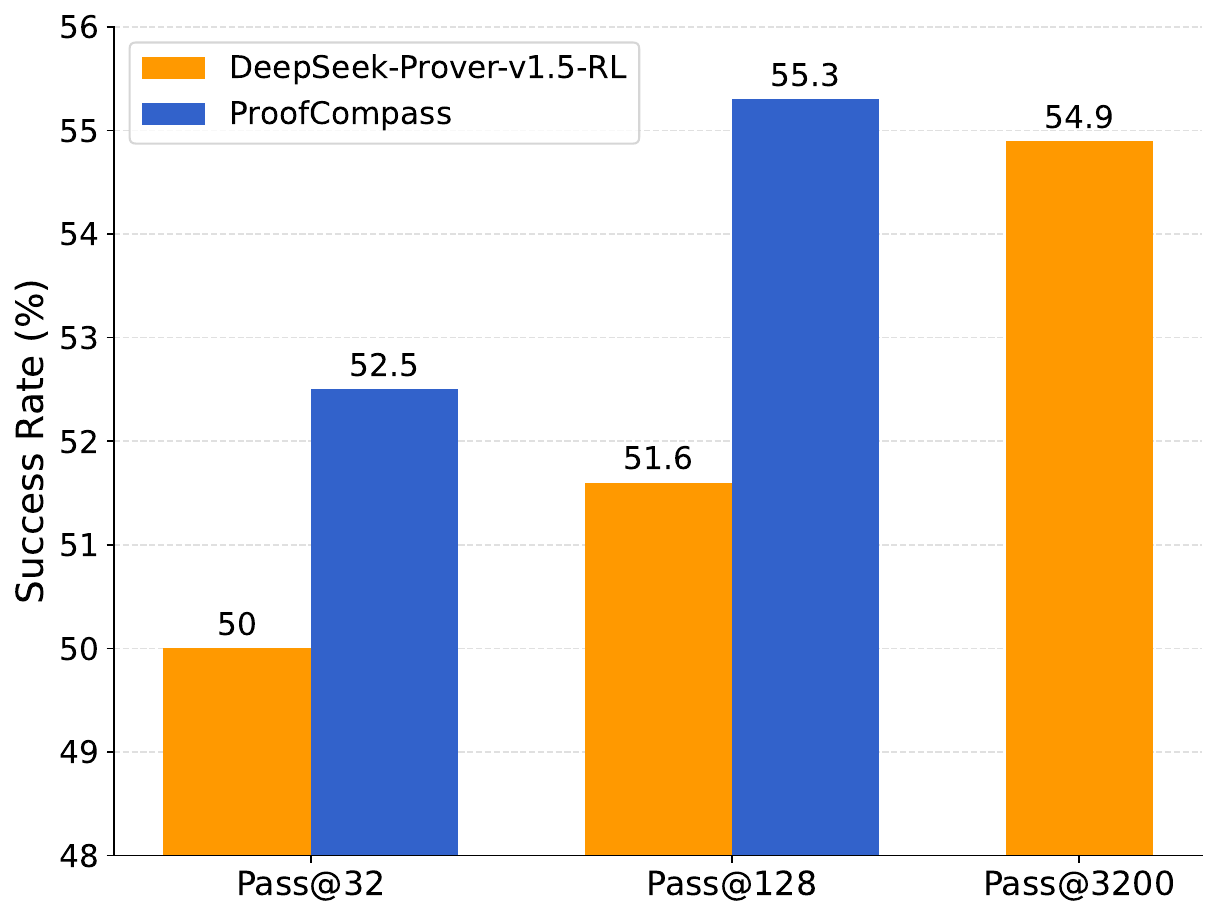}
  \vskip 0.1in
  \caption{Comparison of \systemname{} versus DeepSeek-Prover-v1.5-RL \cite{xin2024deepseek} on the miniF2F-test dataset. Each group of bars shows the side-by-side success rates for a specific Pass@k budget. The figure highlights the resource efficiency of \systemname{}, showing its Pass@128 performance is comparable to the baseline's more computationally expensive Pass@3200 result.}
  \label{fig:table_benchmark}
\end{figure}

The formalization of mathematics, where one seeks to translate mathematical proofs into machine-verifiable form, represents a crucial step toward more reliable mathematical knowledge \cite{avigad2024mathematics}. While proof assistants like Isabelle \cite{Paulson1994-PAUIAG}, HOL Light \cite{harrison-hollight}, and Lean \cite{demoura21lean4} have made formalization possible, the process has historically been labor intensive, requiring deep expertise in both mathematics and formal verification tools. Recent advances in large language models (LLMs) show promise in dramatically changing this landscape, with formal mathematical reasoning emerging as a crucial new frontier \cite{yang2024formal} and prompting broader reflection on the future of mathematical practice \cite{venkatesh2024some,fraser2024will}. For instance, models like GPT-4 \cite{achiam2023gpt} have shown remarkable mathematical reasoning capabilities on complex problems. At the same time, specialized models have achieved unprecedented success in formal mathematics, with the recently released (April 30) DeepSeek-Prover-V2 \cite{ren2025deepseek}, for instance, automatically solving $88.9\%$ of the problems on miniF2F \cite{zheng2021minif2f}, a central benchmark in formal mathematics.

Recent efforts to leverage language models for automated theorem proving have largely followed two distinct paths, each with its own set of strengths and limitations. One path focuses on using large, general-purpose language models, often by employing them within structured, multi-stage algorithms \cite{jiangdraft} that break down the task of formal proof generation. While these large models have shown strong informal mathematical reasoning capabilities, they frequently struggle with the precise syntax and tactics required by proof assistants. Alternatively, a significant body of work has focused on developing specialized models, often at a smaller scale (e.g., 7B parameters or less), tailored specifically for formal mathematics \cite{welleck23llmstep, xin2024deepseek, lin2025goedel, xin2025bfs}. These models excel at generating syntactically correct proofs and closing smaller subgoals, but their relatively modest size inherently limits their mathematical reasoning ability.

More recently, approaches like Kimina-Prover (72B) \cite{wang2025kimina} and DeepSeek-Prover-V2 (671B) \cite{ren2025deepseek} have achieved state-of-the-art results by training large models specifically for theorem proving. However, this paradigm shift towards ever-larger specialized models presents significant challenges. The prohibitive computational resources required for training are largely inaccessible to the broader research community, creating a high barrier to entry and making it difficult to contribute to the advancement of the frontier in AI-assisted theorem proving and formal mathematics.

In this work, we introduce \systemname{}: a synergistic methodology that leverages the broad reasoning capabilities of a large language model to guide a specialized prover, without requiring model training. This approach proves to be highly efficient, matching the average performance of a standalone specialized prover while using a 25-fold smaller computational budget. Furthermore, the framework is designed for modularity: the guiding LLM can be readily swapped, and its core principles are adaptable for use with other specialized provers. By substantially reducing the resources needed to achieve superior results, our work directly addresses the high barrier to entry created by resource-intensive models. This offers a more accessible path forward, helping to democratize cutting-edge research in automated theorem proving.

\begin{figure*}[htbp]
  \centering
  \includegraphics[width=0.7\textwidth]{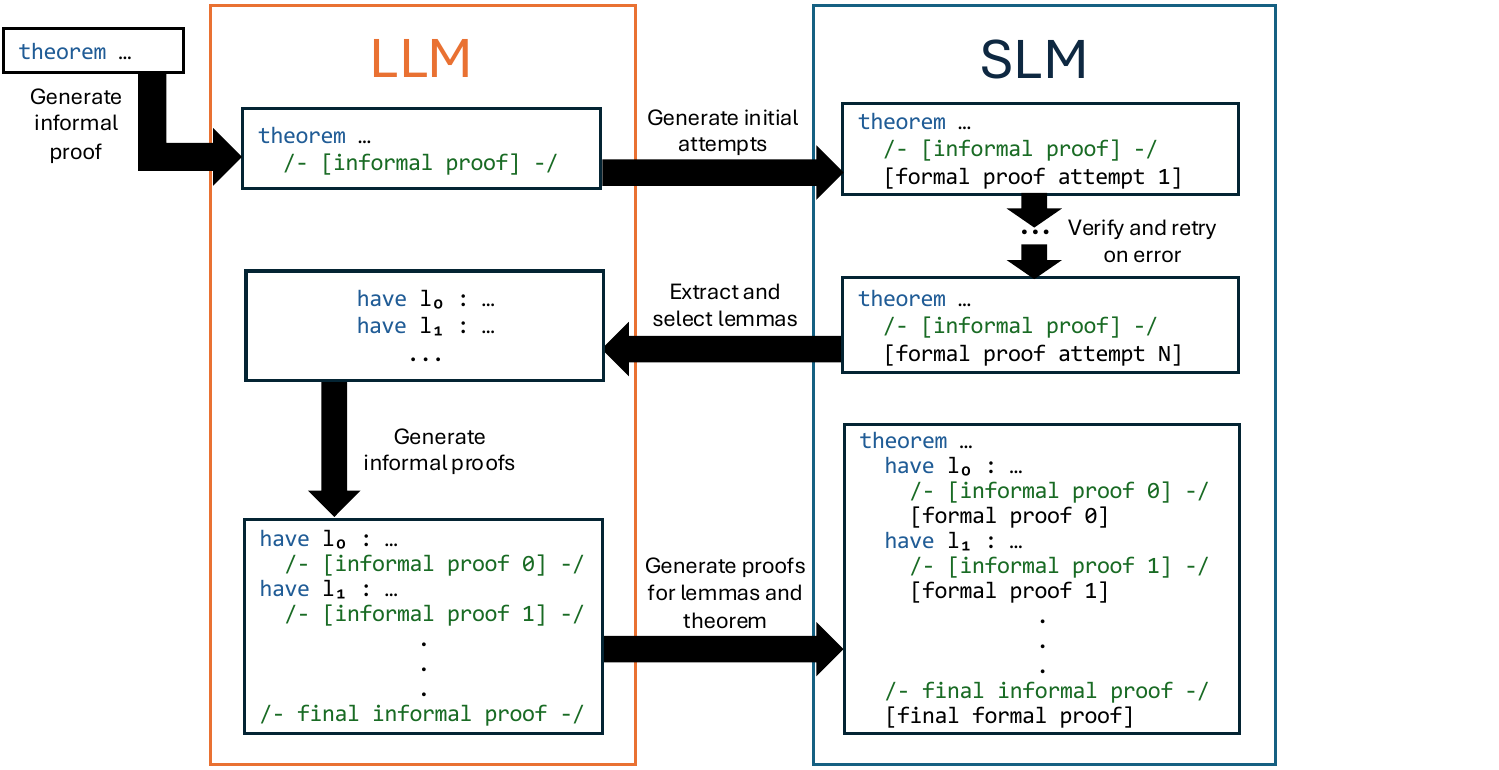}
  \vskip 0.1in
  \caption{Flowchart of \systemname{}. A general-purpose Large Language Model (LLM) first provides natural language guidance for a specialized model (SLM, specifically DSP-v1.5-RL) to iteratively generate and verify proof attempts. If all of these attempts prove unsuccessful, the LLM analyzes the failed attempts to select guiding lemmas, generating further informal proofs to direct the SLM in proving these lemmas and the final theorem.}
  \label{fig:method_flowchart}
\end{figure*}

Concretely, we use a broadly capable LLM to direct the specialized DeepSeek-Prover-v1.5-RL (DSP-v1.5) model \cite{xin2024deepseek}, augmenting its theorem-proving ability in Lean 4 \cite{demoura21lean4} through two primary mechanisms. First, we leverage the LLM's superior natural language understanding and mathematical reasoning abilities to produce a detailed natural language (NL) proof, which, after summarization, is provided to DSP-v1.5 to guide it during formal proof generation. Second, inspired by approaches that utilize intermediate goals or lemmas to structure proof search \cite{wanglego, he2024bc}, we use the LLM to analyze the smaller model's failed attempts. From these attempts, the LLM extracts relevant and correct lemmas, formatted as \texttt{have} statements in Lean. This serves to decompose the original problem into more tractable sub-problems for the specialized prover, creating a structured pathway to solve problems that were initially intractable for the prover working alone. A diagram of the method is shown in Figure~\ref{fig:method_flowchart}.

We evaluate \systemname{} on the miniF2F benchmark \cite{zheng2021minif2f}. The results, displayed in Figure~\ref{fig:table_benchmark}, highlight the powerful efficiency of our synergistic approach. \systemname{} achieves a Pass@128 success rate of $55.3\%$, exceeding the standalone specialized prover's Pass@3200 accuracy of $54.9\%$ while using 25x fewer attempts. The efficiency of our framework is further underscored by its performance on a highly constrained budget: with just 32 attempts, our method's success rate ($52.5\%$) surpasses that of the standalone prover using 4x as many attempts ($51.6\%$ at 128 attempts). These results establish that strategic guidance is a powerfully efficient route to state-of-the-art performance. We summarize our contributions as follows:

\begin{itemize}
    \item We introduce \systemname{}, a flexible framework that synergistically combines a general-purpose LLM with a specialized prover without requiring additional model training. The core approach is broadly portable across different LLMs and specialized provers.
    \item We demonstrate that our method achieves superior performance with considerable resource efficiency on the miniF2F benchmark. Our experiments with DeepSeek-Prover-v1.5-RL show that with a 25-fold smaller budget, our method's Pass@128 accuracy ($55.3\%$) is comparable to the standalone prover's performance with Pass@3200 ($54.9\%$), while Pass@32 ($52.5\%$) also outperforms the prover's Pass@128 ($51.6\%$).
\end{itemize}

\section{Related Work}

Recent advances in automated theorem proving have demonstrated diverse approaches to combining language models with formal reasoning systems. These approaches can be broadly categorized into specialized models designed specifically for formal mathematics, and large, general-purpose language models adapted for theorem proving. Below, we review key developments in each category.

\textbf{Specialized Models.} Holophrasm \cite{whalen2016holophrasm} was one of the earliest works that explored training specialized language models --- then GRUs --- for formal theorem proving, using LMs as policies and value functions in tree search. This foundational effort was later advanced by models like GPT-f \cite{polu2020generative}, which established the viability of using Transformers combined with best-first search. This approach sparked a series of architectural innovations, including Thor \cite{jiang2022thor}, which integrated LLMs and symbolic provers, HTPS \cite{lample2022hypertree}, which introduced a novel tree search algorithm for theorem proving, and ReProver \cite{yang2023leandojo}, which used retrieval-augmented generation. More recently, Lean-STaR \cite{lin24lean} explored training models that interleave informal reasoning in natural language and formal proof steps. POETRY \cite{wangproving} focused on recursive decomposition of proof goals in Isabelle, extracting training data for this decomposition task by restructuring existing human-written proofs. Other models like Goedel-Prover \cite{lin2025goedel} and Mathesis-Prover \cite{xuejun2025mathesis} were trained directly for whole-proof generation, where each pass attempts to generate a complete proof, while InternLM2.5-StepProver \cite{wu2024internlm2}, HunyuanProver \cite{li2024hunyuanprover}, BFS-Prover \cite{xin2025bfs}, \cite{lai2025llm}, and MPS-Prover \cite{liang2025mps} combined specialized models with tree search methods. DeepSeek-Prover-v1.5 \cite{xin2024deepseek} pushed forward both whole-proof generation and tree search.

\textbf{General Language Models.} Another line of work has focused on leveraging general-purpose language models in theorem proving, hoping to benefit from their substantially better performance in a wide range of reasoning tasks. For instance, COPRA \cite{thakur2024an} introduced a model-agnostic language agent framework, initially showing promising results with GPT-4 Turbo \cite{gpt4turbo}. Draft, Sketch and Prove \cite{jiangdraft} used Minerva \cite{lewkowycz2022solving} to first write a natural language proof and then sketch it as an Isabelle proof with subgoals to be proved by hammers. Other recent work leveraging general-purpose LLMs includes Lyra \cite{zhenglyra}, Subgoal-based prover \cite{zhao2023decomposing}, and LEGO-Prover \cite{wanglego}.

Alternatively, automated proof repair frameworks such as ProofAug \cite{liu2025efficient} and APOLLO \cite{ospanov2025apollo} employ programmatic harnesses with compiler feedback and built-in solvers to correct failed attempts. In contrast, we leverage a second LLM for analyzing failures. Unlike works that use an LLM to directly attempt to fix proofs, like Baldur \cite{first2023baldur}, we use an LLM to extract correct lemmas using the outcomes of failed attempts.

\textbf{Hybrid Approaches.} Recognizing the complementary strengths of specialized and general-purpose models, a promising research direction involves their synergistic combination. For instance, BC-Prover \cite{he2024bc} employs an LLM for backward chaining and sub-goal generation, and delegating tactic-level proof steps to specialized provers. In contrast, our method leverages an LLM to analyze the failed proof attempts of a specialized prover to select useful intermediate lemmas, rather than having the LLM directly generate sub-goals.

\textbf{Specialized large models.} Apart from hybrid systems, another powerful yet resource-intensive approach involves directly specializing large models such as Kimina-Prover (72B) \cite{wang2025kimina} and DeepSeek-Prover-V2 (671B) \cite{ren2025deepseek} through further training, a strategy that yields state-of-the-art results but at an often prohibitive cost at most compute budgets. Our work, in contrast, focuses on synergistically combining existing models without incurring these significant training overheads, aiming for a more resource-efficient path to improved performance.

\section{Method}

In this section, we describe our proposed method for automated theorem proving through the synergistic guidance of a specialized prover by a general-purpose LLM.

\subsection{Preliminaries}
\label{preliminaries}

\begin{figure*}[htbp]
  \centering
  \includegraphics[width=0.75\textwidth]{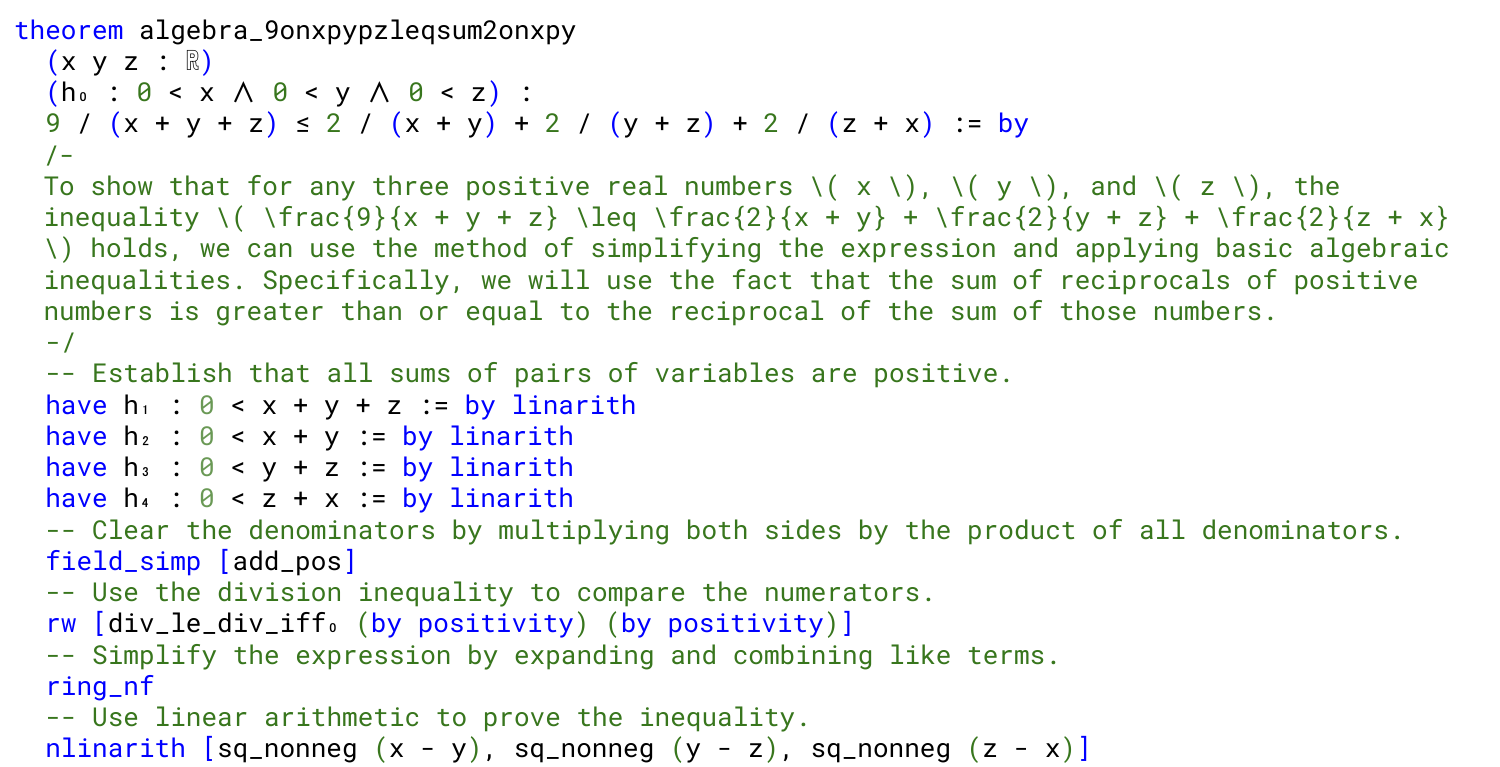}
  \vskip 0.1in
  \caption{Sample proof by DSP-v1.5 of a theorem in Lean 4.}
  \label{fig:sample_proof}
\end{figure*}

Our methodology builds upon the DeepSeek-Prover-v1.5-RL (DSP-v1.5) model \cite{xin2024deepseek}, which exhibits a distinct proof generation structure. DSP-v1.5 operates in two primary modes: a Chain-of-Thought (CoT) \cite{wei2022chain} mode and a non-CoT mode. These can be activated using specific prompts, as detailed in \cite{xin2024deepseek}.

The CoT mode initiates the formal proof by generating an informal, natural-language proof sketch (Figure~\ref{fig:sample_proof}). This sketch is embedded within Lean's block comment syntax (`/- ... -/') and is placed immediately after the theorem statement. Following this informal sketch, DSP-v1.5 proceeds to generate the formal Lean 4 code, often interleaving it with short explanatory comments prefixed with `--'. In contrast, the non-CoT mode of DSP-v1.5 generates only Lean 4 code, without any accompanying informal proofs or comments. In this work, we focus on the CoT mode of DSP-v1.5. A common characteristic of DSP-v1.5's proof generations is the production of intermediate steps articulated as `have' statements within the Lean proof. These `have' statements effectively function as lemmas or sub-goals, and are a common mechanism for structuring proofs in Lean.

These distinct operational characteristics of DSP-v1.5 inform our approach by suggesting two main avenues for external guidance, which our method aims to exploit:
\begin{itemize}
    \item \textbf{Informal Proof Guidance:} The initial informal proof section generated by DSP-v1.5 in CoT mode provides a natural interface for supplying a high-level, natural language (NL) proof strategy. We leverage this by employing a larger, more capable LLM whose superior reasoning abilities can generate more effective informal proofs than DSP-v1.5 might produce autonomously. This approach is motivated by findings in works such as \cite{xin2024deepseek} and \cite{lin24lean}, which demonstrated performance improvements by training models on Lean proofs annotated with natural language reasoning steps, thereby showing that NL reasoning can enhance specialized models' performance in formalization tasks. Works like \cite{jiangdraft, wanglego, zhenglyra} also use LLM-generated proofs, which they use to guide the general model in its proof generation.
    \item \textbf{Lemma-based Guidance:} The `have' statements offer a mechanism for decomposing the main theorem and guiding the proof search via structured intermediate goals. An LLM can analyze failed proofs from DSP-v1.5 and strategically select lemmas that effectively partition the problem. Each such lemma then represents a simpler, more constrained subproblem for DSP-v1.5. This strategy of using lemmas or subgoals to augment theorem proving has proven effective in prior systems like BC-Prover \cite{he2024bc} and LEGO-Prover \cite{wanglego}, which utilize them to structure and simplify the proof search process. Moreover, works such as Lyra \cite{zhenglyra} and ProofAug \cite{liu2025efficient} achieve significant performance gains by repairing failed proofs using automated tools like Sledgehammer \cite{PAAR-2010:Three_Years_Experience_with} and predefined sets of tactics, thereby demonstrating the potential to extract valuable components from initially unsuccessful derivations.
\end{itemize}

Our framework’s architecture is designed for broad applicability. The guiding LLM is fully interchangeable, allowing the system to immediately benefit from future model improvements, while the core lemma-selection mechanism is adaptable to any specialized prover that generates structured proof attempts. The only component tailored to specific prover features is the injection of informal proof texts, which leverages the CoT-style output of DSP-v1.5 in our experiments. Consequently, \systemname{} is directly compatible with models that share the exact generation structure of DSP-v1.5, such as Goedel-Prover \cite{lin2025goedel}. For other provers, the informal proof injection must be modified to suit different CoT structures \cite{wang2025kimina} or can be omitted entirely for those lacking CoT support \cite{xin2025bfs}, with the core lemma-based guidance remaining fully effective in either case.

We now detail how these guidance mechanisms are implemented. Section \ref{informal_proof} presents our informal proof guidance approach, where the LLM enhances the theorem-proving capabilities of the specialized prover (SLM, DSP-v1.5 in our experiments), while Section \ref{lemma_extraction} describes the lemma-based guidance mechanism

% We now detail how the aforementioned guidance channels are leveraged, beginning in the next section with our approach to informal proof guidance by the LLM to enhance the theorem-proving capabilities of the Specialized Language Model (SLM), which in this paper is DSP-v1.5.

\subsection{Informal Proof Generation and Initial Attempts}
\label{informal_proof}

The first stage of \systemname{} focuses on providing the SLM with a high-quality, natural language (NL) proof to guide its CoT mode.

To achieve this, we initially prompt an LLM with the formal theorem statement $T_{formal}$ and its informal counterpart $T_{informal}$\footnote{In miniF2F \cite{zheng2021minif2f}, formal statements are already paired with natural language counterparts. When only a formal statement is available, the LLM itself could be used to \emph{informalize} the statement if needed.} to generate a complete natural language proof, denoted as $p_{NL}$:
\begin{equation}
    (T_{formal}, T_{informal}) \xrightarrow{\text{LLM}} p_{NL} \label{eq:gen_pnl}
\end{equation}

The full NL proof $p_{NL}$ often exceeds the input context capacity of the SLM. Thus, we employ a second LLM query to summarize it, yielding a concise proof summary $p_{summary}$:
\begin{equation}
    (T_{formal}, T_{informal}, p_{NL}) \xrightarrow{\text{LLM}} p_{summary} \label{eq:gen_summary}
\end{equation}
$p_{summary}$ is then embedded within the Lean block comment syntax (`/- ... -/') expected by DSP-v1.5's CoT mode, appended to the formal theorem statement.

With this LLM-generated proof summary ($p_{summary}$) integrated into the input, we query DSP-v1.5 (using its CoT activation prompt) to sequentially generate and verify individual proof attempts. This process continues until either a correct proof is found or the maximum of $N$ initial attempts is reached. Each attempt $i \in \{1, \dots, N\}$ corresponds to a single generation call:

\begin{equation}
    (T_{formal}, p_{NL}) \xrightarrow{\text{SLM}} P_{init}^i \label{eq:gen_init}
\end{equation}

Each attempt consumes one unit of the total budget $B$. If a correct proof is found, the process concludes successfully, and the subsequent lemma extraction and proving stages are skipped. If the $N$-attempt limit is reached without success, the method proceeds as detailed in the following sections.

\subsection{LLM-Guided Lemma Selection}
\label{lemma_extraction}

Should the initial $N$ proof attempts fail to prove the theorem, \systemname{} proceeds to extract and select a set of guiding lemmas. This process begins by aggregating all unique `have' statements extracted from $P_{init}^1, \dots, P_{init}^{N}$ generated by the SLM. This yields a candidate pool of potential lemmas, $L_{extract}$:
\begin{equation}
    P_{init}^1, \dots, P_{init}^{N} \xrightarrow{\text{lemma extraction}} L_{extract}
\end{equation}

The candidate pool $L_{extract}$ is first subjected to an automated syntax check to filter out any malformed statements. This initial validation step ensures that only syntactically correct lemmas are passed to the LLM for evaluation, resulting in the set of valid lemmas, $L_{valid}$:
\begin{equation}
    L_{extract} \xrightarrow{\text{syntax checking}} L_{\text{valid}}
\end{equation}

This set of valid lemmas is then processed by the guiding LLM. Given the pool $L_{valid}$, the formal theorem statement $T_{formal}$, its informal counterpart $T_{informal}$, the complete LLM-generated natural language proof $p_{NL}$ for the theorem, and the maximum number of lemmas to select $k$, the LLM is tasked with selecting at most $k$ lemmas, $L_{select} = (l_0, l_1, \dots, l_{m-1}), m \leq k$:
\begin{equation}
    (L_{valid}, T_{formal}, T_{informal}, p_{NL}, k) \xrightarrow{\text{LLM}} L_{select} \label{eq:llm_selection}
\end{equation}

To produce $L_{select}$, the LLM analyzes each candidate lemma in $L_{valid}$. This involves evaluating the lemma's correctness, primarily considering whether the statement is (i) logically sound and provable using only the global hypotheses of $T_{formal}$, and (ii) directly justified by $p_{NL}$. From the subset of lemmas deemed correct, the LLM then performs a strategic selection. This selection aims to identify lemmas covering key inferential steps of $p_{NL}$, choosing at most $k$ lemmas to constitute the final output set $L_{select}$, which are ordered according to the logical flow of $p_{NL}$. The rationale for this approach of extracting and selecting lemmas from SLM attempts, as opposed to direct LLM generation, is detailed in Appendix \ref{alternative_lemmas}.

In the second stage, the lemmas in $L_{select}$ are passed to the LLM for informal proof generation. The LLM generates a concise informal proof $p_l$ for each selected formal lemma $l \in L_{select}$, derived from $p_{NL}$, $T_{formal}$ and $T_{informal}$. It also generates a concise informal proof for the main theorem, $p_{main}$, which assumes the truth of all selected lemmas in $L_{select}$:
\begin{align}
    &(l, T_{formal}, T_{informal}, p_{NL}) \xrightarrow{\text{LLM}} p_l \\
    &(T_{formal}, T_{informal}, L_{select}, p_{NL}) \xrightarrow{\text{LLM}} p_{main}
\end{align}
The LLM adapts explanations from $p_{NL}$ to generate a proof for each lemma $l$. The collection of all these generated informal proofs is denoted by $\mathcal{P} = \{ p_l \mid l \in L_{select} \} \cup \{ p_{main} \}$.

Next, $L_{proven}$ is initialized by extracting lemma proofs from previous attempts. For each $l \in L_{select}$, we search $P_{init}^1, \dots, P_{init}^{N}$ for all its occurrences as a `have' statement. For each such occurrence with an associated proof $P_{l,extract}$, we verify if $P_{l,extract}$ proves $l$ using only the global hypotheses of $T_{formal}$. If this check passes and $l$ is not already in $L_{proven}$, we add $l$ and that $P_{l,extract}$ to $L_{proven}$:
\begin{equation}
    (P_{init}^1, \dots, P_{init}^{N}, T_{formal}, L_{select}) \rightarrow L_{proven} \label{eq:salvage_lemmas}
\end{equation}

The outcome of this stage is thus the curated set of formal lemmas $L_{select}$, the set of their corresponding LLM-generated informal proofs $\mathcal{P}$, and the initialized set of proven lemmas $L_{proven}$. These are then used to guide subsequent formal proof attempts by DSP-v1.5, as detailed in the next section.

\subsection{Lemma Proving and Theorem Validation}
\label{lemma proving}

This stage proceeds to prove the theorem $T_{formal}$ using the set of selected lemmas $L_{select}$ and their corresponding informal proofs $\mathcal{P}$. In case no lemmas were selected, the method reverts to the initial proof generation strategy described in Section~\ref{informal_proof} for the remainder of the total budget $B$.

Throughout this stage, each call to the SLM to generate $P_{main}$ or any individual lemma proof $P_{l_i}$ consumes an attempt from the overall budget $B$ allocated for the problem. If the attempts allocated for this lemma-based strategy are exhausted before a complete and verified proof $P_{final}$ is obtained, this approach is considered unsuccessful for the current theorem.

Initially, the method seeks to obtain $P_{main}$, a formal proof of the main theorem $T_{formal}$ under the assumption that all lemmas in $L_{select}$ hold. The SLM is prompted to generate $P_{main}$ using $T_{formal}$ (with all lemmas in $L_{select}$ appended as hypotheses) and the LLM-generated informal proof for the main theorem $p_{main}$:
\begin{equation}
    (T_{formal}, L_{select}, p_{main}) \xrightarrow{\text{SLM}} P_{main} 
\end{equation}
This $P_{main}$ provides the concluding logic to derive $T_{formal}$ from the established lemmas.

If $P_{main}$ is successfully generated, an iterative procedure begins. Within each iteration of this loop, the method first attempts to prove individual lemmas. It iterates through each lemma $l_i \in L_{select}$ (for $i$ from $0$ to $m-1$). If $l_i$ is not already in $L_{proven}$, $l_i$ is formulated as a distinct Lean 4 theorem. This theorem includes all global hypotheses of $T_{formal}$ and all previously ordered lemmas from $L_{select}$ that precede $l_i$ as hypotheses, with the goal being the formal statement of $l_i$. The SLM then attempts to generate a formal proof $P_{l_i}$ for this lemma, guided by $l_i$'s informal proof $p_{l_i}$:
\begin{equation}
    (l_0, ..., l_i, T_{formal}, p_{l_i}) \xrightarrow{\text{SLM}} P_{l_i}
\end{equation}
If $P_{l_i}$ is successfully generated and verified by Lean 4, $l_i$ is added to $L_{proven}$.

Following this lemma-proving pass within an iteration, a candidate full proof, $P_{final}$, is constructed for $T_{formal}$. This $P_{final}$ starts with $T_{formal}$ and its global hypotheses, incorporates `have' statements for each lemma $l_k \in L_{proven}$ along with their formal proofs $P_{l_k}$, and finally appends the proof steps from the previously generated $P_{main}$. This complete $P_{final}$ is then submitted to Lean 4 for validation. If $P_{final}$ is a valid proof of $T_{formal}$, the theorem is considered proven, and the process concludes successfully.

\section{Experiment}
\label{experiment}

In this section we describe the experimental evaluation of our proposed method.

\subsection{Experimental Setup}

\textbf{Dataset.} We evaluate our approach on the widely used miniF2F dataset \cite{zheng2021minif2f}. It contains 488 formal mathematics problems, consisting of olympiad-level problems (AMC, AIME, IMO) and undergraduate mathematics exercises formalized in various proof assistant systems, including Isabelle \cite{Paulson1994-PAUIAG}, HOL Light \cite{harrison-hollight}, Metamath \cite{metamath}, and Lean \cite{demoura21lean4}. The Lean problems were originally in Lean 3, but were converted to Lean 4 in \cite{yang2023minif2f}.

\begin{table}[t]
\label{sample-table}
\caption{Pass@k performance on miniF2F-test}
\vskip 0.1in
\begin{center}
\begin{small}
\begin{sc}
\begin{tabular}{lcccr}
\toprule
Method & Budget & miniF2F-test \\
\midrule
Baselines \\
\midrule
BC-Prover & 100 & $30.7\%$ \\
BC-Prover + ReProver & 100 & $31.6\%$ \\
BC-Prover + LLMStep & 100 & $32.0\%$ \\
LEGO-Prover & 100 & $50.0\%$ \\
Lyra & 200 & $51.2\%$ \\
DSP-v1.5 & 32 & $50\%\pm0.5\%$ \\
& 128 & $51.6\%\pm0.5\%$ \\
& 3200 & $54.9\%\pm0.7\%$ \\
\midrule
Ours \\
\midrule
\systemname{} & 32 & $52.5\%$ \\
& 128 & $\textbf{55.3\%}$ \\
\midrule
Ablations \\
\midrule
Only Lemma Guidance & 32 & $51.6\%$ \\
& 128 & $53.3\%$ \\
Only Informal Proof & 32 & $52.0\%$ \\
& 128 & $52.9\%$ \\
\bottomrule
\end{tabular}
\end{sc}
\end{small}
\end{center}
%\vskip -0.1in
\label{main_results}
\end{table}

\textbf{Evaluation.} We assess the performance of \systemname{}  using the Pass@k metric, specifically reporting Pass@32 and Pass@128 due to computational budget constraints. For our methodology, an ``attempt'' in the Pass@k calculation corresponds to any call made to the DSP-v1.5 model. This includes initial proof generation attempts, as well as individual attempts to prove each selected lemma and the theorem itself using the lemmas (see Appendix \ref{llm_overhead} for a detailed breakdown of LLM task and SLM attempt timings, including overhead considerations). The validity of generated formal proofs and selected lemmas is verified by interacting with Lean 4 via the Lean REPL \cite{leanprover_community_repl}. A proof attempt is deemed successful if the Lean REPL returns no error messages and the generated proof does not contain any \texttt{sorry} placeholders. Conversely, an attempt is considered a failure if it produces errors, includes \texttt{sorry} placeholders, or if the Lean verification process exceeds a 20-second timeout.

\textbf{Baselines.} Our primary point of comparison is the performance of the DSP-v1.5 model as reported in its original publication \cite{xin2024deepseek}. To contextualize our method's performance, we also compare against BC-Prover \cite{he2024bc}, a relevant LLM-driven framework for theorem proving in Lean (considered both as a standalone LLM-based prover and in its hybrid configurations with ReProver \cite{yang2023leandojo} and LLMStep \cite{welleck23llmstep}), and other prominent LLM-based provers such as LEGO-Prover \cite{wanglego} and Lyra \cite{zhenglyra}, all of which utilize the Isabelle proof assistant. This selection of baselines reflects our study's central aim: to introduce and evaluate a novel LLM-guidance methodology for a given specialized prover, using DSP-v1.5 as our specific test case. We therefore do not include comparisons with other standalone specialized provers, as our primary goal is to assess the LLM-guidance technique itself, rather than to conduct a wide-ranging benchmark of all prover types.

\textbf{Implementation Details.} Our methodology utilizes specific Large Language Models for its guidance components. The initial, comprehensive natural language proof ($p_{NL}$) for a given theorem is generated using Gemini 2.0 Flash Thinking Experimental \cite{Kane2025}. All other LLM-driven tasks within our method, such as summarizing $p_{NL}$, performing lemma analysis and selection, and elaborating the proofs for each lemma, are performed using Gemini 2.0 Flash \cite{PichaiHassabisKavukcuoglu2024}. For all interactions with the DeepSeek-Prover-v1.5-RL model, we employ its Chain-of-Thought (CoT) mode, using the specific prompt provided in \cite{xin2024deepseek}. All attempts were generated using an Nvidia GeForce RTX 4090, and verification was conducted using Lean version 4.15. 

\textbf{Hyperparameters.} As we compare our method with the Pass@32 and Pass@128 performances of DSP-v1.5, the total generation budget is $B=128$. The initial number of attempts was chosen to be $N=16$, as this often resolves a significant portion of problems upfront by DSP-v1.5 and provides a sufficient corpus of `have' statements from failed attempts for effective LLM guidance, while selecting at most $k = 5$ lemmas focuses on key strategic steps. Appendix \ref{hyperparameters} elaborates on these choices and includes supporting analysis.

\subsection{Main Results}

The performance of our proposed method on the miniF2F-test dataset is presented in Table \ref{main_results}, with a visual comparison to the DeepSeek-Prover-v1.5-RL baseline for Pass@32 and Pass@128 rates depicted in Figure \ref{fig:table_benchmark}. Our approach demonstrates notable improvements in theorem-proving capabilities and resource efficiency, particularly when enhancing its foundational specialized prover, DSP-v1.5.

When compared directly to this baseline, our method yields substantial gains. As shown in Table \ref{main_results}, our Pass@32 rate of $52.5\%$ exceeds the baseline's Pass@128 rate of $51.6\%$. More strikingly, with a 25-fold smaller computational budget, our method's Pass@128 rate of $55.3\%$ matches the average $54.9\%$ performance that DSP-v1.5 achieves with 3200 attempts.

Further contextualizing these results, our method also shows strong performance relative to other contemporary systems. A key contribution of this work is its novel hybrid architecture. To the best of our knowledge, BC-Prover \cite{he2024bc} is the most comparable contemporary hybrid approach; our method (55.3\% Pass@128) significantly outperforms BC-Prover + LLMStep ($32.0\%$ Pass@100), underscoring the efficacy of our specific synergistic design. Moreover, even with a limited budget of 32 attempts, our method's Pass@32 rate of $52.5\%$ is competitive with or exceeds other LLM-based methods using $3-6\times$ larger attempt budgets, such as LEGO-Prover ($50.0\%$ at 100 attempts) and Lyra ($51.2\%$ at 200 attempts).

These combined results underscore the significant advantage of our synergistic approach, achieving strong performance and resource efficiency by effectively leveraging the complementary strengths of general-purpose LLMs and specialized provers.

\subsection{Ablation Study}
\label{ablation}

To better understand the contributions of the primary components of our methodology (LLM-generated informal proof guidance and LLM-guided lemma selection) we conducted two ablation studies. The results are included in Table \ref{main_results}.

The first ablation, ``Only Lemma Guidance,'' isolates the impact of the lemma-based guidance mechanism. In this configuration, the initial LLM-generated informal proof $p_{summary}$, subsequent informal proofs for lemmas $p_{l_i}$ and the main theorem $p_{main}$ are omitted. Instead, DSP-v1.5 generates its own informal proofs (as in its standard CoT mode, illustrated in Figure \ref{fig:sample_proof}), while still benefiting from the LLM's analysis of failed attempts to extract and select guiding `have' statements (lemmas). This ablation proved highly effective on its own; in particular, its Pass@32 performance of $51.6\%$ already matches the baseline's Pass@128 result, and its total Pass@128 rate reached $53.3\%$. This suggests that strategically decomposing the problem using LLM-selected lemmas already aids the specialized prover, even without direct informal proof guidance from the LLM for each step. However, a high-quality informal proof still provides an extra performance boost.

The second ablation, ``Only Informal Proof,'' evaluates the effectiveness of providing LLM-generated informal proofs alone without the subsequent lemma processing. Here, the LLM generates the initial informal proof $p_{summary}$, which guides DSP-v1.5 for all $N=128$ initial attempts (as described in Section \ref{informal_proof}), but the lemma extraction, selection, and proving stages (Sections \ref{lemma_extraction} and \ref{lemma proving}) are entirely skipped. This configuration yielded a Pass@32 of $52.0\%$ and a Pass@128 of $52.9\%$, which also outperforms the baseline. Notably, its Pass@32 rate surpasses the baseline's Pass@128 rate ($51.6\%$), demonstrating that leveraging the LLM's advanced reasoning for high-level proof strategies enables the specialized prover to find correct formalizations with significantly fewer attempts by effectively narrowing the search space.

Together, these results highlight that both the LLM-generated informal proof guidance and the LLM-driven lemma processing contribute to the overall performance of our method. Their contributions are complementary: their combination yields the best performance. \systemname{} thus demonstrates the benefits of a synergistic interaction between the LLM and the specialized prover.

\section{Conclusion}

In this paper, we introduced \systemname{}, a novel methodology to enhance automated theorem proving by synergistically leveraging the high-level reasoning of general-purpose Large Language Models (LLMs) to guide specialized provers. Our approach employs an LLM to generate strategic natural language proofs and to analyze failed proof attempts by a specialized model to extract and select useful intermediate lemmas, thereby decomposing complex problems.

The empirical validation on the miniF2F \cite{zheng2021minif2f} benchmark highlights the framework's remarkable resource efficiency. \systemname{} is on par with the 3200-attempt average performance of the standalone prover while using only 128 attempts---a 25-fold computational saving. This efficiency extends to smaller scales, where our Pass@32 performance surpasses the baseline's Pass@128 accuracy. This work thus establishes that intelligently orchestrating the complementary strengths of LLMs and specialized provers offers a more promising and resource-efficient path towards more capable theorem proving systems.

\section{Limitations and Future Work}

The proposed methodology, while demonstrating promising results, presents several limitations that also outline avenues for future research.

While our core lemma selection loop is designed for broad applicability to any prover generating structured proof attempts, the informal proof guidance component in its current form is tailored to the specific CoT-style input of provers like DSP-v1.5. A key avenue for future work is therefore to both validate the core lemma-based framework across a more diverse range of specialized provers and to develop more generalized techniques for the informal guidance component. 

Furthermore, the experimental evaluation presented herein was conducted with a maximum of $2^{7}$ attempts per problem. This represents a considerably more constrained computational budget compared to several contemporary works in automated theorem proving, which often report Pass@k metrics frequently exceeding $2^{17}$ attempts \cite{li2024hunyuanprover, wu2024internlm2, xin2024deepseek}. Consequently, an important avenue for future research involves assessing the performance scaling of our methodology under significantly increased computational resources.

\newpage

\section*{Acknowledgements}

This study was financed in part by the Coordenação de Aperfeiçoamento de Pessoal de Nível Superior - Brasil (CAPES) -- Finance Code 001. The authors also wish to thank Professor Adriano Côrtes for his valuable assistance with the computational experiments in this work.

\section*{Impact Statement}

This work presents a resource-efficient methodology for automated theorem proving. By demonstrating a path forward that does not require large-scale computational infrastructure, our approach helps to democratize research in this domain. We believe this focus on computational efficiency is a positive contribution that can foster broader and more inclusive participation within the machine learning community.


\bibliography{citations}
\bibliographystyle{icml2025}

%%%%%%%%%%%%%%%%%%%%%%%%%%%%%%%%%%%%%%%%%%%%%%%%%%%%%%%%%%%%%%%%%%%%%%%%%%%%%%%
%%%%%%%%%%%%%%%%%%%%%%%%%%%%%%%%%%%%%%%%%%%%%%%%%%%%%%%%%%%%%%%%%%%%%%%%%%%%%%%
% APPENDIX
%%%%%%%%%%%%%%%%%%%%%%%%%%%%%%%%%%%%%%%%%%%%%%%%%%%%%%%%%%%%%%%%%%%%%%%%%%%%%%%
%%%%%%%%%%%%%%%%%%%%%%%%%%%%%%%%%%%%%%%%%%%%%%%%%%%%%%%%%%%%%%%%%%%%%%%%%%%%%%%
\newpage
\appendix
\onecolumn

\begin{center}
    \textbf{ Supplementary material for\\ \Large\systemname{}: Enhancing Specialized Provers with LLM Guidance}
\end{center}

This supplementary document is organized as follows:

\begin{itemize}
  \item \textbf{Section~\ref{llm_overhead}} presents a computational time analysis for each component of the proposed workflow.
  \item \textbf{Section~\ref{hyperparameters}} provides justification for key hyperparameter choices.
  \item \textbf{Section~\ref{alternative_lemmas}} compares the chosen strategy of extracting lemmas from the specialized prover's failed attempts with direct LLM generation.
  \item \textbf{Section~\ref{llm_prompts}} provides the complete prompts used for all LLM tasks to ensure reproducibility.
\end{itemize}

\section{Generation Times}
\label{llm_overhead}

% Figure environment with two minipages - Assuming this LaTeX code for figures is correct and provided by you.
\begin{figure}[htbp]
    \centering
    \includegraphics[width=0.7\textwidth]{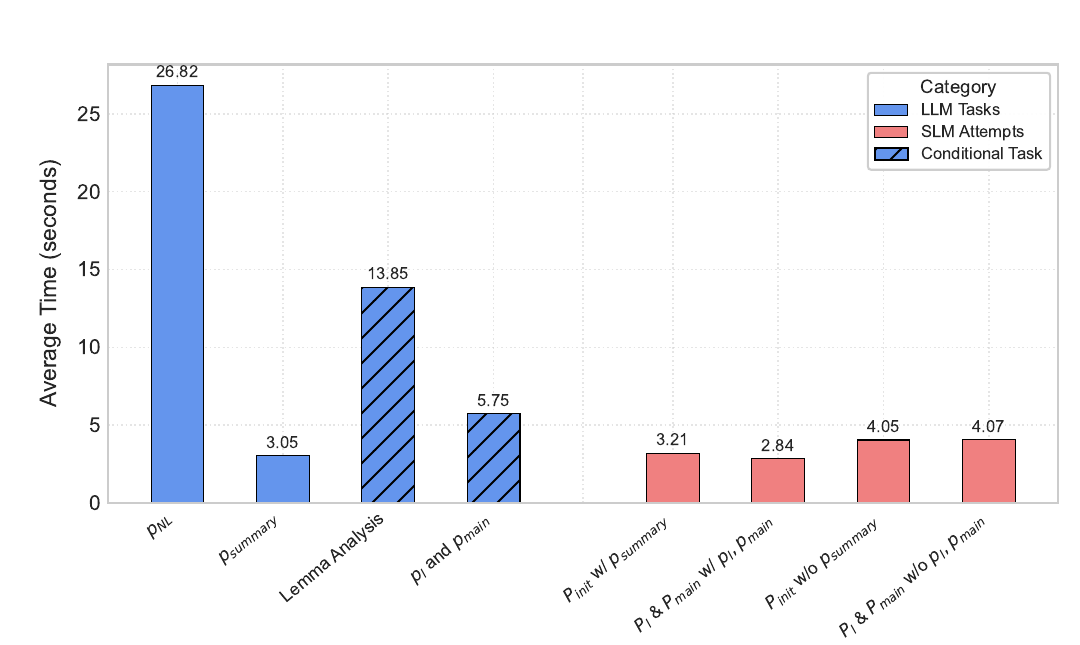}
    \vskip 0.1in
    \caption{A detailed breakdown of the average time, in seconds, for each component of the \systemname{} workflow compared to the baseline. The chart categorizes operations into three types: one-time LLM guidance tasks (solid blue), conditional LLM tasks for lemma processing (striped blue), and per-attempt costs for the Specialized Language Model (SLM) (red).}
    \label{fig:average_generation_times}
\end{figure}

This section details the computational timings of individual LLM guidance tasks and Specialized Language Model (SLM) proof attempts, followed by an analysis of their impact on overall time-to-solution performance.

Figure~\ref{fig:average_generation_times} presents the average durations for key operations. Initial LLM tasks, such as $p_{NL}$ generation and subsequent summarization, constitute an upfront time investment, with specific durations detailed in the figure. Lemma analysis also introduces an overhead if a problem is not solved within N = 16 initial attempts. A crucial benefit of this LLM pre-processing is evident in the SLM attempt times: proof attempts guided by an LLM-generated $p_{summary}$ are notably faster (average 3.21s) than the baseline DSP-v1.5 CoT attempts (average 4.05s). This speed-up occurs because the baseline model, in its CoT mode, must first generate its own informal proof sketch, as illustrated in Figure~\ref{fig:sample_proof}, whereas our method provides this sketch as input. Similarly, the generation of informal proofs for lemmas $(p_{l})$ and the main theorem $(p_{main})$, while adding to LLM task time, leads to more efficient SLM generation for the corresponding lemma and final theorem proofs.

A quantitative analysis reveals when the upfront time cost of our method is offset by its greater per-attempt efficiency. For problems requiring more than $N=16$ attempts, \systemname{} incurs a significant fixed time overhead of approximately 100.83 seconds. This accounts for initial LLM guidance tasks (29.87s), the first 16 failed SLM attempts (51.36s), and the conditional lemma processing (19.60s). Following this, each subsequent attempt to prove a lemma or the main theorem is highly efficient, costing only 2.84 seconds. In contrast, the baseline method costs 4.05 seconds per attempt, regardless of the stage. Consequently, the baseline's cumulative time surpasses \systemname{}'s full workflow after approximately 46 total attempts. This indicates that while the baseline may be faster for problems solved in fewer than 46 attempts, \systemname{}'s approach becomes more time-efficient for more complex problems requiring a larger computational budget.

\section{Hyperparameters Analysis}
\label{hyperparameters}

\begin{figure*}[htbp]
  \centering
  \includegraphics[width=0.7\textwidth]{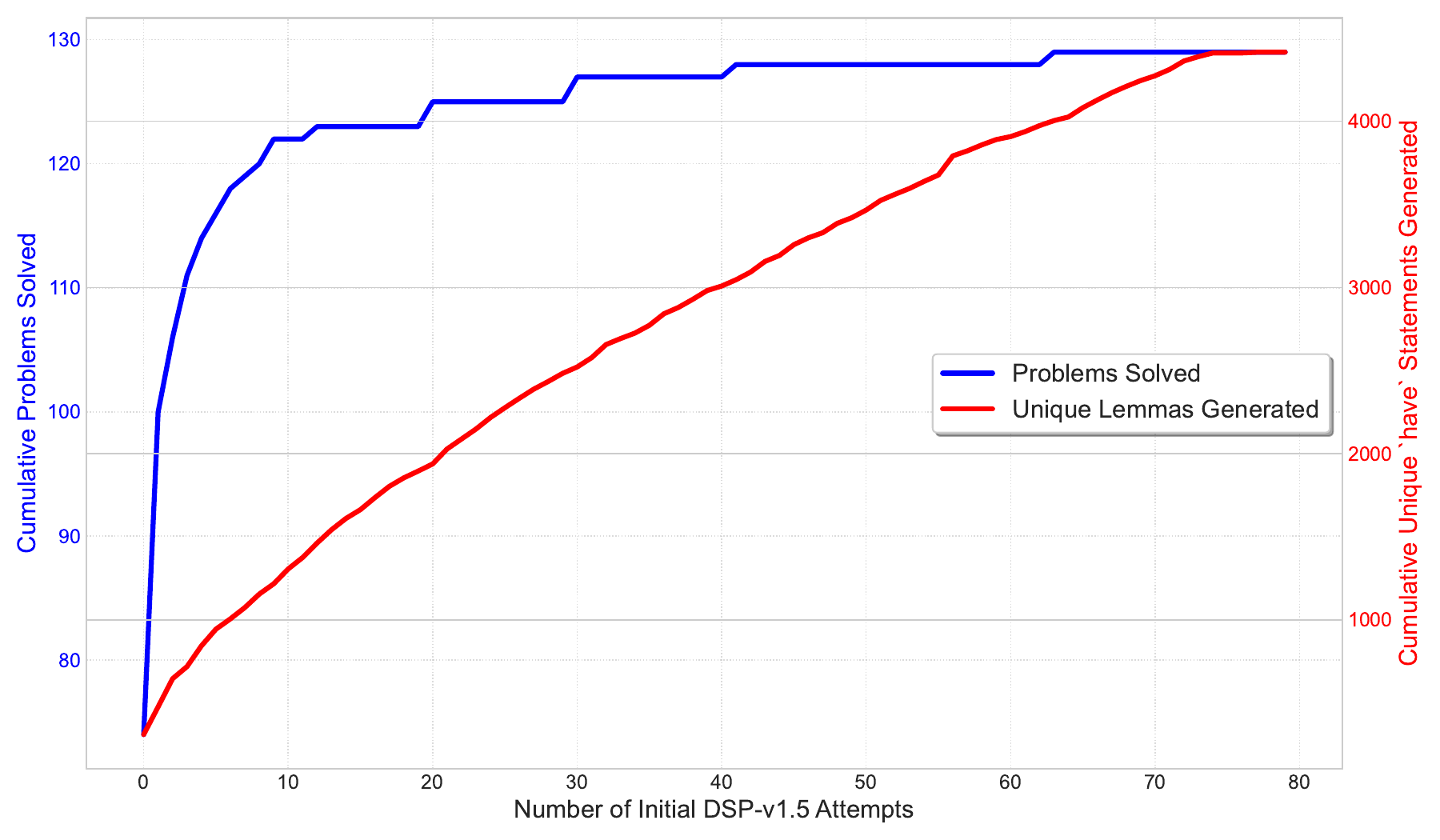}
  \vskip 0.1in
  \caption{Impact of the number of initial DSP-v1.5 attempts ($N$) on the cumulative number of problems solved (blue line, left y-axis) and the cumulative number of unique `have' statements (potential lemmas) generated (red line, right y-axis). This data was collected from the initial proof generation stage (Section~\ref{informal_proof}), where DSP-v1.5 was guided by the LLM-generated proof summary, prior to any subsequent lemma extraction (Section~\ref{lemma_extraction}) or proving (Section~\ref{lemma proving}) cycles.}
  \label{fig:hyperparameters}
\end{figure*}

This section provides the justification for the key hyperparameters used in the experiments (Section~\ref{experiment}). The number of initial proof attempts by DSP-v1.5 was set to $N=16$. This choice is informed by the trends observed in Figure~\ref{fig:hyperparameters}. While the rate of solving new problems (\textcolor{blue}{blue curve}) significantly diminishes after approximately $N=10-15$ attempts, the pool of unique `have' statements (potential lemmas, \textcolor{red}{red curve}) continues to expand, reaching around 2000 unique statements close to $N=20$. Setting $N=16$ thus aims to capture a majority of the problems solvable by DSP-v1.5 in its initial attempts while still generating a diverse corpus of material from failed attempts for effective LLM-guided lemma extraction (Section~\ref{lemma_extraction}).

Regarding the maximum number of selected lemmas, this was set to $k=5$. This value was established through preliminary qualitative evaluations, which indicated that up to five strategically chosen lemmas were generally adequate to cover the essential inferential steps of most proofs. Our methodology also allows the LLM to select fewer than $k$ lemmas if a smaller set is deemed optimal. Requesting a larger $k$ in these evaluations risked the inclusion of trivial or redundant lemmas, which could dilute the specialized prover's focus and inefficiently expend its attempt budget on less critical sub-problems. Therefore, $k=5$ encourages the LLM to identify a concise yet impactful set of guiding lemmas for DSP-v1.5.

\section{Alternative Lemma Generation Strategies}
\label{alternative_lemmas}

\begin{figure}[htbp]
    \centering
    \begin{minipage}[t]{0.48\linewidth}
        \centering
        \includegraphics[width=0.95\textwidth]{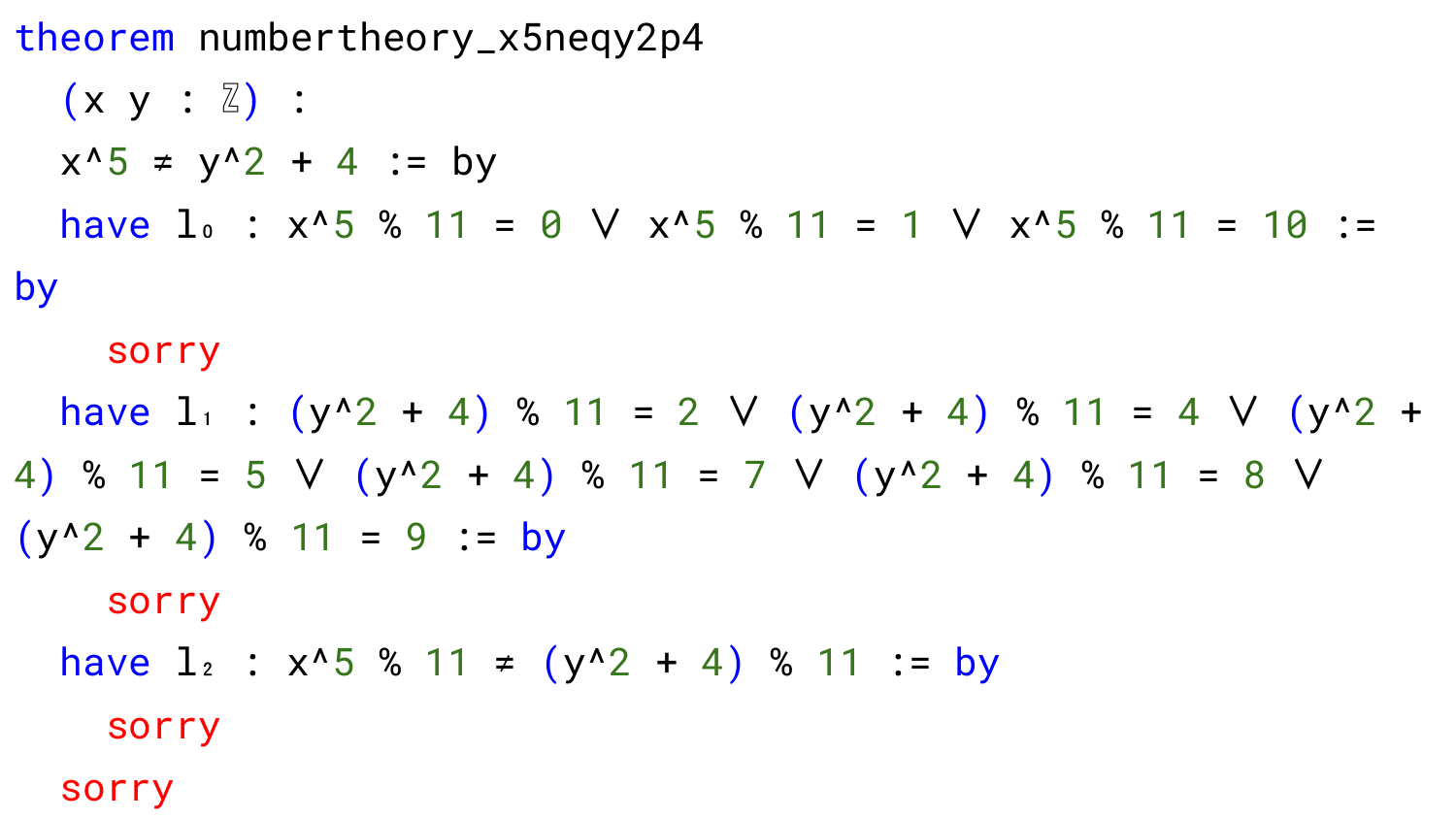}
        \vskip 0.1in
        \caption{Lemmas extracted from failed SLM attempts.}
        \label{fig:lemmas_slm}
    \end{minipage}%
    \hfill
    \begin{minipage}[t]{0.48\linewidth}
        \centering
        \includegraphics[width=0.95\textwidth]{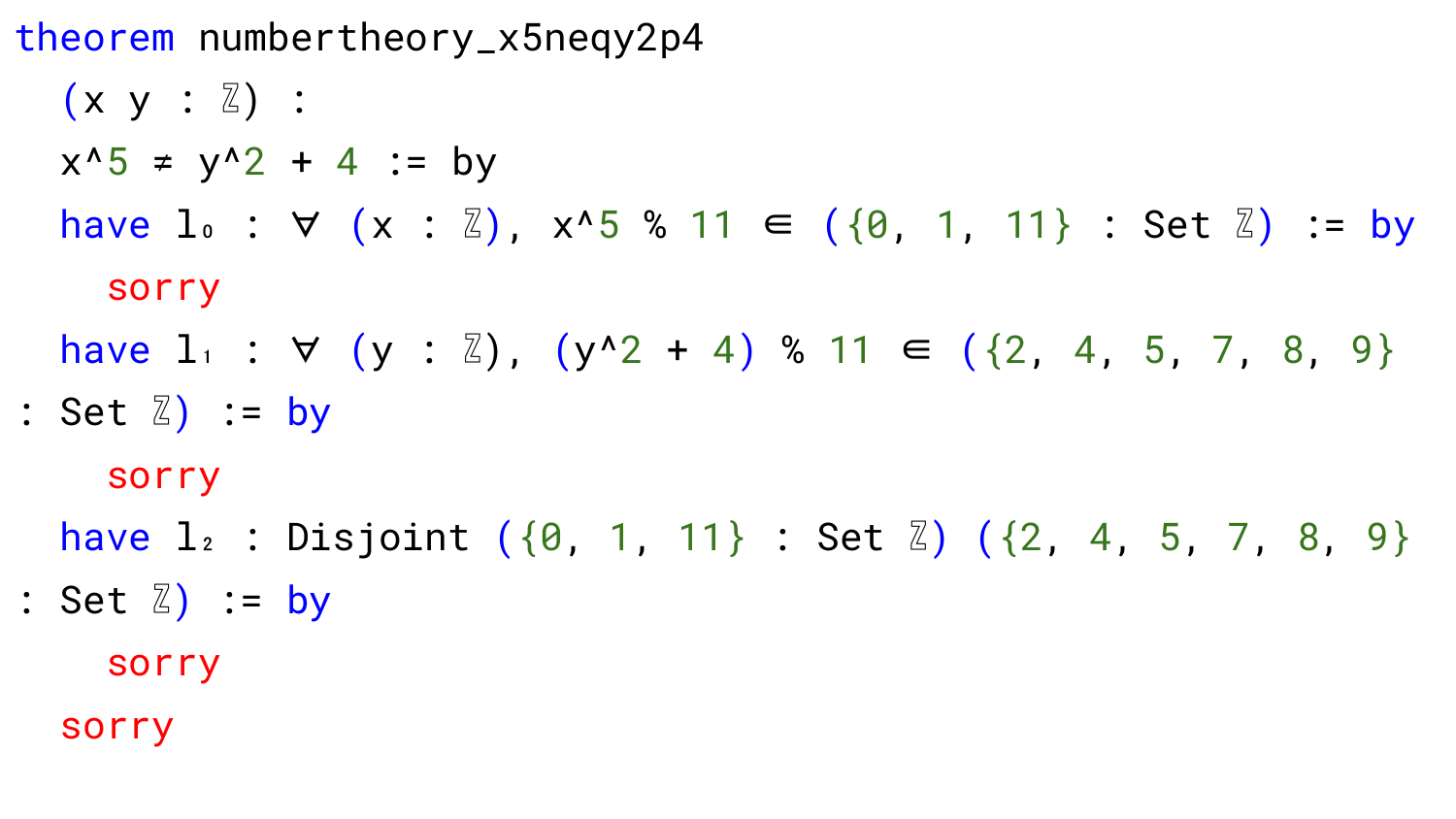}
        \vskip 0.1in
        \caption{Lemmas generated by the LLM}
        \label{fig:lemmas_llm}
    \end{minipage}
\end{figure}

An alternative to our adopted method of extracting and selecting lemmas from the Specialized Language Model's (SLM, DSP-v1.5) attempts is to directly prompt a Large Language Model (LLM) to generate intermediate lemmas. This section elaborates on why the former approach was chosen. The primary reason, identified in early experimentation, is that lemmas directly generated by an LLM, while often semantically correct, tended to be syntactically different from those typically produced by DSP-v1.5 and proved more challenging for the SLM to utilize.

For instance, in the problem \texttt{numbertheory\_x5neqy2p4}, directly LLM-generated lemmas often express constraints using set theory (e.g., stating a variable is an element of a set of possible values), as can be seen in Figure~\ref{fig:lemmas_llm}. In contrast, lemmas extracted from SLM attempts (Figure~\ref{fig:lemmas_slm}) frequently articulate equivalent constraints as disjunctions of direct equality assertions for the variable. While semantically equivalent, these syntactic variations significantly impacted the SLM's proving efficiency. In a comparative experiment where the SLM was tasked with proving theorems using these two styles of guiding lemmas (requiring proofs for each lemma $P_{l_i}$ and the main theorem $P_{main}$), the approach using LLM-extracted, SLM-style lemmas required an average of 12 attempts. Conversely, using directly LLM-generated lemmas required an average of 17.25 attempts (both averaged over 4 independent runs), a difference of over 40\%.

This empirical difference suggests that the SLM is more adept at processing lemmas that conform to syntactic patterns closer to its own typical outputs or training data. Therefore, our methodology prioritizes extracting and selecting lemmas from the SLM's own attempts. This aim is to provide strategically sound intermediate steps in a form that is more tractable for the SLM, thereby optimizing the overall resource efficiency of the theorem-proving process.

\section{LLM Prompts}
\label{llm_prompts}

This appendix provides all prompts used in our experiments for reproducibility.

\subsection{User Prompt for Natural Language Proof Generation}

\begin{tcolorbox}[
  breakable,
  colback=white,
  colframe=black,
  arc=3mm,
  fonttitle=\bfseries
]
Provide a proof in natural language for the theorem below:

Theorem statement in Lean 4:

$<$formal\_statement$>$

Informal Statement:

$<$informal\_statement$>$

Please provide a rigorous, detailed proof using natural language.
\end{tcolorbox}

\subsection{System Prompt for Proof Summarization}

\begin{tcolorbox}[
  breakable,
  colback=white,
  colframe=black,
  arc=3mm,
  fonttitle=\bfseries
]
You are a mathematical proof summarizer. You will be given a formal statement of a theorem in Lean 4, its informal counterpart, and a natural language proof of the theorem. Your response should contain only a summarized version of the natural language proof that's both concise and complete. You should start you response with ``We want to show that", ``We have", ``We need to show that" or ``To show that". You should write only the proof in your response, nothing more.
\end{tcolorbox}

\subsection{User Prompt for Proof Summarization}

\begin{tcolorbox}[
  breakable,
  colback=white,
  colframe=black,
  arc=3mm,
  fonttitle=\bfseries
]
THEOREM STATEMENT IN LEAN 4:

$<$formal\_statement$>$

INFORMAL STATEMENT:

$<$informal\_statement$>$

COMPLETE PROOF:

$<$nl\_proof$>$

Write your summarized proof below
\end{tcolorbox}

\subsection{System Prompt for Lemma Selection}

\begin{tcolorbox}[
  breakable,
  colback=white,
  colframe=black,
  arc=3mm,
  fonttitle=\bfseries
]
You are a theorem-proving assistant trained to evaluate individual statements for their provability as standalone lemmas in Lean 4. Your task is to analyze a single list of candidate lemma statements, and then select at most 5 to cover all key steps of the natural language proof.

\textbf{Core Principle for Lemma Evaluation}

A candidate lemma statement $L$ is ``correct'' if and only if a Lean 4 statement \texttt{have [name] : [statement] := by [proof]} would be successfully provable using ONLY the global assumptions/hypotheses provided in the formal theorem statement.

A lemma is ``incorrect'' if it couldn't be proven in Lean 4 using only the assumptions and hypothesis provided in the, or if it's not fully justified by the natural language proof based on the global assumptions.

\textbf{Specific Criteria Guiding ``Correct'' vs. ``Incorrect'' Evaluation}

\begin{enumerate}
    \item \textbf{Global Provability:} The \textbf{exact mathematical statement} of the lemma must be provably true using only the given hypotheses in the formal theorem statement.

    \item \textbf{Justification by NL Proof:} The lemma must be a direct and logical step or assertion found in the natural language proof, understandable from the global context.

    \item \textbf{No Dependence on Undischarged Assumptions:}
    \begin{itemize}
        \item If the proof proceeds by cases (e.g., ``Case 1: Assume $P$... then $R$'', ``Case 2: Assume $Q$... then $T$''), a lemma stating $P$ by itself, $Q$ by itself, $R$ by itself (if $R$ depends on $P$), or $T$ by itself (if $T$ depends on $Q$) is \textbf{incorrect}. These statements are only true under temporary, local assumptions.
        \item However, a lemma stating $P \lor Q$ (if this disjunction is provable globally) would be \textbf{correct}.
        \item Similarly, $P \rightarrow R$ or $Q \rightarrow T$ might be \textbf{correct} if these implications are what the proof establishes.
    \end{itemize}

    \item \textbf{Contradictions \& Unjustified Steps:}
    \begin{itemize}
        \item If a lemma contradicts a statement in the proof or a hypothesis, it's \textbf{incorrect}.
        \item If a lemma makes an assertion not present or derivable from the NL proof and global hypotheses, it's \textbf{incorrect}.
        \item In a proof by contradiction, if `$A$' is true and the proof temporarily assumes `not $A$' to reach a contradiction, a lemma stating `not $A$' as a factual step from the global context is \textbf{incorrect}.
    \end{itemize}
    
    \item \textbf{Consequences of Incorrect Lemmas:} If a lemma $B$ follows logically from another lemma $A$, and lemma $A$ is determined to be ``incorrect'', then lemma $B$ is also ``incorrect''.
\end{enumerate}

\textbf{Input Format}
You'll receive the formal statement of the theorem in Lean 4, its informal counterpart, a complete natural language proof, and a list of candidate lemma statements.

\textbf{THEOREM STATEMENT IN LEAN 4:}\\
$<$Formal statement in Lean 4$>$

\textbf{INFORMAL STATEMENT:}\\
$<$Informal translation of theorem statement$>$

\textbf{COMPLETE PROOF:}\\
$<$Complete natural language proof of theorem$>$

\textbf{LEMMAS:}\\
0: $<$lemma in Lean 4$>$\\
1: $<$lemma in Lean 4$>$\\
$...$ \\
N-1: $<$lemma in Lean 4$>$

\textbf{Output Format (Strictly Adhere to This Structure)}
\textbf{LEMMA ANALYSIS}\\
0:\\
Analysis: [Provide a short, step-by-step analysis of lemma 0. Explain precisely why it IS or IS NOT provable as a standalone `have' statement given ONLY the theorem's global hypotheses and the NL proof. Reference the \textbf{Core Principle} and \textbf{Specific Criteria} above.]\\
Evaluation: [correct or incorrect]

1:\\
Analysis: [Analysis for lemma 1 as above.]\\
Evaluation: [correct or incorrect]\\
... 

N-1:\\
Analysis: [Analysis for lemma N-1 as above.]\\
Evaluation: [correct or incorrect]

\textbf{SELECTION RATIONALE FOR CHOSEN LEMMAS}\\
\textbf{AVAILABLE LEMMAS:} [List all lemmas that you evaluated as 'correct'.]\\
\textbf{REASONING:} [Based on your 'Evaluation' of all original lemmas above, select up to 5 correct lemmas that represent the key intermediate steps from the provided natural language proof. Do not select lemmas that are restatements of the theorem's hypotheses or its final conclusion. Always try to select 5 lemmas, choosing fewer only if there are not enough correct intermediate lemmas available. The final chosen lemmas must be ordered to reflect the logical flow of the natural language proof.]

\textbf{CHOSEN LEMMAS}\\

[Write all chosen lemmas here. Each chosen lemma must be presented in the following format. Number them sequentially starting from $l_0$, regardless of their original index.]\\

have $l_0$ : $<$statement$>$ := by\\
have $l_1$ : $<$statement$>$ := by\\
...

\textbf{This is how your output should look like}

\textbf{LEMMA ANALYSIS}\\
0:\\
Analysis: [Analysis of lemma 0 based on the criteria explained above]\\
Evaluation: [Evaluation of lemma 1 (correct or incorrect)]

1:\\
Analysis: [Analysis of lemma 0 based on the criteria explained above]\\
Evaluation: [Evaluation of lemma 1 (correct or incorrect)]
$...$ 

N-1:\\
Analysis: [Analysis of lemma 0 based on the criteria explained above]\\
Evaluation: [Evaluation of lemma 1 (correct or incorrect)]

\textbf{SELECTION RATIONALE FOR CHOSEN LEMMAS}\\
\textbf{AVAILABLE LEMMAS:}

[first correct lemma statement]

$...$

[last correct lemma statement]

\textbf{REASONING:} [Reasoning about which (AT MOST 5) lemmas to choose, with the goal of selecting the key intermediate steps of the NL proof, excluding hypotheses and the conclusion, and ordering them according to the proof's logical flow.]

\textbf{CHOSEN LEMMAS:}

[List of chosen lemmas. Note that the subscripts should be in the order 0, 1, 2, ..., irrespective of the number of its corresponding lemma in the input.]\\

have $l_0$ : $<$statement$>$ := by\\
have $l_1$ : $<$statement$>$ := by\\

\end{tcolorbox}

\subsection{User Prompt for Lemma Selection}

\begin{tcolorbox}[
  breakable,
  colback=white,
  colframe=black,
  arc=3mm,
  fonttitle=\bfseries
]
\textbf{THEOREM STATEMENT IN LEAN 4:}\\
$<$formal\_statement$>$

\textbf{INFORMAL STATEMENT:}\\
$<$informal\_statement$>$

\textbf{COMPLETE PROOF:}\\
$<$nl\_proof$>$

\textbf{LEMMAS:}\\
0: $<$lemma 0$>$\\
1: $<$lemma 1$>$\\
...\\
N-1: $<$lemma N-1$>$

Proceed with your response below, strictly following all guidelines and output structures specified in the system prompt.

\end{tcolorbox}

\subsection{System Prompt for Lemma and Final Proof Generation}

\begin{tcolorbox}[
  breakable,
  colback=white,
  colframe=black,
  arc=3mm,
  fonttitle=\bfseries
]
You are a mathematical proof analyst. Your task is to analyze a given mathematical proof (provided alongside formal and informal statements) and break it down into a sequence of logical steps, formatted clearly for potential formalization in Lean 4.

You will receive:
\begin{enumerate}
    \item A formal theorem statement in Lean 4, potentially including hypothesis labels like $h_0$, $h_1$.
    \item An informal statement of the theorem.
    \item A complete natural language proof of the theorem.
    \item Formal lemma statements (e.g., \textbf{have} $l_0$ : $<...>$ ) that will appear in the Lean proof.
\end{enumerate}

Analyze the provided natural language proof and follow these guidelines meticulously to structure your output:

\begin{enumerate}
    \item \textbf{REASONING Section:}
    \begin{itemize}
        \item Start with \textbf{REASONING:}.
        \item Analyze the provided natural language proof to identify its overall strategy and key insights.
        \item Explicitly note how the formal lemma statements (\textbf{have} $l_0$, \textbf{have} $l_1$, etc.) map to the key steps in the natural language proof. These will directly correspond 1-to-1 with the $l_0:$, $l_1:$ steps you generate. You must NEVER include new lemma statements that are not present in the input.
        \item Briefly list the main steps you identified in the provided proof, ensuring each step aligns with one of the provided formal lemma statements. Remember to never include new lemma statements in the steps.
        \item In the case the formal lemma statements don't match the steps in the proof well, you should try to modify the natural language proof to make it match the formal lemma statements. You should NEVER modify the formal lemma statements themselves, or add new lemma statements.
    \end{itemize}

    \item \textbf{STEPS Section:}
    \begin{itemize}
        \item Follow with \textbf{STEPS:}.
        \item Label the key milestones as $l_0:$, $l_1:$, etc. These must exactly match the order and content of the formal \textbf{have} statements provided.
        \item Each step ($l_i:$) must state a precise mathematical fact in natural mathematical language.
    \end{itemize}

    \item \textbf{Step Proofs (\textbf{Proof:} sections):}
    \begin{itemize}
        \item Follow each $l_i:$ statement with \textbf{Proof:}.
        \item Provide a concise summary of the justification found in the provided natural language proof.
        \item To show dependency on a previous step $l_j$, state the mathematical result of $l_j$ \textit{as a fact} within the narrative summary of the proof for $l_i$, instead of mentioning the name $l_j$ directly. You should NEVER mention the name $l_j$ inside the proof.
        \item Correct: ``Since $<$mathematical statement from $l_j$ holds / was established$>$, the proof proceeds by..."
        \item Incorrect: ``By $l_j$..." or ``Using the result from $l_j$...". This is incorrect because '$l_j$' is present in the proof, which is not allowed.
    \end{itemize}
    
    \item \textbf{Final Proof Section:}
    \begin{itemize}
        \item End with \textbf{Final Proof:}.
        \item Summarize how the proof combines the results stated in the $l_i$ steps to reach the final conclusion.
        \item State intermediate results directly as established facts, without referencing step labels.
    \end{itemize}

    \item \textbf{Formatting:}
    
    Your response must follow this exact structure:

    \textbf{REASONING:}\\
    $<$Analysis noting correspondence between formal lemmas and steps$>$

    \textbf{STEPS:}\\
    $l_0:$\\
    $<$First mathematical statement matching first \textbf{have}$>$\\
    \textbf{Proof:}\\
    $<$Detailed explanation with all necessary calculations$>$

    $l_1:$\\
    $<$Next mathematical statement matching next \textbf{have}$>$\\
    \textbf{Proof:}\\
    $<$Detailed explanation with all necessary calculations$>$

    ...

    \textbf{Final Proof:}\\
    $<$Conclusion using established results$>$
\end{enumerate}

\textbf{Here's an example showing the expected output format:}

$<$example from miniF2F-valid$>$

\end{tcolorbox}

\subsection{User Prompt for Lemma and Final Proof Generation}

\begin{tcolorbox}[
  breakable,
  colback=white,
  colframe=black,
  arc=3mm,
  fonttitle=\bfseries
]
\textbf{THEOREM STATEMENT IN LEAN 4:}\\
$<$formal\_statement$>$

\textbf{INFORMAL STATEMENT:}\\
$<$informal\_statement$>$

\textbf{COMPLETE PROOF:}\\
$<$nl\_proof$>$

\textbf{FORMAL LEMMAS::}\\
$<$lemmas$>$
\end{tcolorbox}

\end{document}